\documentclass[10pt,twocolumn,letterpaper]{article}

\usepackage{color}
\usepackage{epsfig}
\usepackage{graphicx}

\usepackage{adjustbox}
\usepackage{array}
\usepackage{booktabs}
\usepackage{colortbl}
\usepackage{wrapfig}
\usepackage{hhline}
\usepackage{multirow}
\usepackage{wrapfig}
\usepackage{floatflt}

\usepackage{amsmath,amsfonts,amssymb}
\usepackage{mathtools}  %
\usepackage{bm}
\usepackage{nicefrac}
\usepackage{microtype}

\usepackage{changepage}
\usepackage{extramarks}
\usepackage{fancyhdr}
\usepackage{lastpage}
\usepackage{setspace}
\usepackage{soul}
\usepackage{xspace}

\usepackage{enumerate}
\usepackage{enumitem}  %

\usepackage{makecell}

\usepackage{pifont} %

\usepackage{algorithm,algpseudocode}

\usepackage{amsthm}

\usepackage[symbol]{footmisc}
\newcolumntype{L}[1]{>{\raggedright\let\newline\\\arraybackslash\hspace{0pt}}m{#1}}
\newcolumntype{C}[1]{>{\centering\let\newline\\\arraybackslash\hspace{0pt}}m{#1}}
\newcolumntype{R}[1]{>{\raggedleft\let\newline\\\arraybackslash\hspace{0pt}}m{#1}}

\newcommand{\ignore}[1]{}

\makeatletter
\DeclareRobustCommand\onedot{\futurelet\@let@token\@onedot}
\def\@onedot{\ifx\@let@token.\else.\null\fi\xspace}

\def\eg{e.g\onedot}

\makeatother

\definecolor{MyDarkBlue}{rgb}{0,0.08,0.8}
\definecolor{MyDarkGreen}{RGB}{45,155,45}
\definecolor{MyDarkRed}{rgb}{0.8,0.02,0.02}
\definecolor{MyOrange}{rgb}{1.0, 0.4, 0.2}
\definecolor{MyPurple}{RGB}{111,0,255}
\definecolor{MyRed}{rgb}{0.8,0.0,0.0}
\definecolor{MyGold}{rgb}{0.75,0.6,0.12}
\definecolor{MyDarkgray}{rgb}{0.66, 0.66, 0.66}
\definecolor{JiayuanColor}{rgb}{0.60,0.43,0.48}

\newcommand{\model}{LARC\xspace}

\newcommand{\cmark}{\ding{51}}%
\newcommand{\xmark}{\ding{55}}%

\usepackage{cvpr}              %
\usepackage[dvipsnames]{xcolor}

\definecolor{cvprblue}{rgb}{0.21,0.49,0.74}
\usepackage[pagebackref,breaklinks,colorlinks,citecolor=cvprblue]{hyperref}

\def\paperID{2403} %
\def\confName{CVPR}
\def\confYear{2024}

\title{Naturally Supervised 3D Visual Grounding with \\ Language-Regularized Concept Learners}

\author{Chun Feng\thanks{Equal contribution.}\\
Stanford University\\
\and
Joy Hsu\footnotemark[1]\\
Stanford University\\
\and
Weiyu Liu\\
Stanford University\\
\and
Jiajun Wu\\
Stanford University\\
}

\begin{document}
\maketitle
\begin{abstract}
3D visual grounding is a challenging task that often requires direct and dense supervision, notably the semantic label for each object in the scene. In this paper, we instead study the naturally supervised setting that learns from only 3D scene and QA pairs, where prior works underperform. We propose the Language-Regularized Concept Learner (\model), which uses constraints from language as regularization to significantly improve the accuracy of neuro-symbolic concept learners in the naturally supervised setting. Our approach is based on two core insights: the first is that language constraints (\eg, a word's relation to another) can serve as effective regularization for structured representations in neuro-symbolic models; the second is that we can query large language models to distill such constraints from language properties. We show that \model improves performance of prior works in naturally supervised 3D visual grounding, and demonstrates a wide range of 3D visual reasoning capabilities---from zero-shot composition, to data efficiency and transferability. Our method represents a promising step towards regularizing structured visual reasoning frameworks with language-based priors, for learning in settings without dense supervision. 
\end{abstract} %
\vspace{-0.3cm}
\section{Introduction}
\begin{figure}[tp!]
  \centering
  \includegraphics[width=\linewidth]{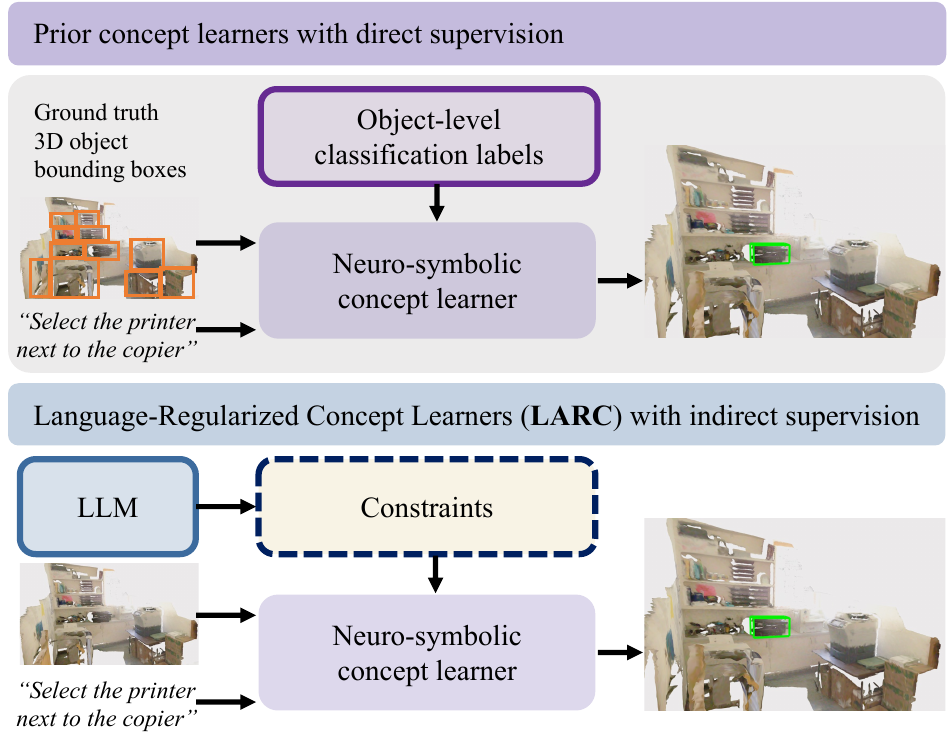}
  \vspace{-0.6cm}
  \caption{Compared to prior works, \model conducts 3D visual grounding in the naturally supervised setting, by training neuro-symbolic concept learners with language regularization.}
\label{fig:pull}
\vspace{-0.5cm}
\end{figure}

3D visual reasoning models often require direct supervision during training to achieve faithful 3D visual grounding, for example, in the form of classification labels for each ground truth object bounding box in the scene. However, dense visual annotation for 3D scenes is difficult and expensive to acquire. %
In this paper, we study the more practical setting of \textit{naturally supervised} 3D visual grounding, where models learn by looking at only scene and question-answer pairs, and do not use object-level classification supervision. Prior state-of-the-art works for 3D referring expression comprehension \cite{jain2022bottom, huang2022multi, bakr2022look, dailan2018transref, junha2021lanref, hsu2023ns3d} do not show strong visual reasoning capabilities, such as generalization and data efficiency, in this indirectly supervised setup (See Figure~\ref{fig:pull}).

To this end, we propose a neuro-symbolic concept learner that leverages constraints from language (\eg, a word's relation to another), as regularization in low guidance settings. Compared to visual annotations, language-based priors are cheap to annotate, and free when distilled from large language models (LLMs). We show that especially in settings with indirect supervision, models benefit from such regularization as it reduces overfitting on noisy signals. Notably, we leverage the structured and interpretable representations from neuro-symbolic methods to effectively inject constraint-based regularization.

Neuro-symbolic concept learners decompose visual reasoning queries into modular functions, and execute them with neural networks. Each network outputs representations that can be indexed by its concept name (\eg, \textit{chair}) and arity (\eg, that the concept describes a \textit{binary} relation) \cite{NSCL}. Due to this modularity, prior neuro-symbolic works have shown strong generalization, data efficiency, and transferability in the 3D domain \cite{hsu2023ns3d}. However, they require direct classification supervision to train intermediate networks for concept grounding, and do not take into account the taxonomy of the concept name in relation to other concepts in language, nor its effect on other concepts. %
For example, previous methods do not capture from language that \textit{wardrobe} and \textit{dresser} are visually similar and synonymous, or that a vase \textit{left} of a painting indicates that the painting is \textit{right} of the vase. Although recent neuro-symbolic works have used LLMs for the interpretation of language to function structures \cite{hsu2023ns3d, wong2023word, hsu2023s}, they purely translate the query and do not encode language priors in execution.

To enable learning in naturally supervised settings, we propose a \textbf{La}nguage-\textbf{R}egularized \textbf{C}oncept Learner (\textbf{\model}). \model builds upon the 3D neuro-symbolic concept learner, NS3D \cite{hsu2023ns3d}, and introduces regularization on intermediate representations based on language constraints. Our method leverages general constraints derived from well-studied semantic relations between words, for example, symmetry~\cite{gleitman1996similar}, exclusivity~\cite{markman1988children}, and synonymity~\cite{miller1995wordnet}, which are broadly applicable across all language-driven tasks.
We show that we can effectively encode such semantic contexts from language into neuro-symbolic concept learners through regularization. Additionally, we demonstrate that concepts satisfying these constraints can be distilled from LLMs based on language priors (\eg, the concepts \textit{near} and \textit{far} are symmetric), and that language rules also enable execution of novel concepts from composition of learned concepts.%

\model significantly improves performance of neuro-symbolic concept learners in two datasets for 3D referring expression comprehension. In addition, we demonstrate that using language-based rules allows \model to zero-shot execute unseen concepts, while all previous models fail to generalize. Importantly, \model significantly improves data efficiency and transferability between datasets compared to prior works, critical for 3D visual reasoning systems, while not requiring object-level classification labels. %

In summary, our key contributions are the following:
\begin{itemize}[noitemsep]
\item We propose language-based constraints that capture the semantic properties of concepts, and show that such priors can be distilled from large language models. %
\item We introduce constraint-based regularization that can be effectively applied on the structured intermediate representations of neuro-symbolic concept learners.
\item We show that we can leverage language rules to compose unseen concepts from learned concepts.
\item We empirically validate \model's improvement upon prior works on two datasets, and show strong generalization, data efficiency, and transferability between datasets.
\end{itemize} %
\section{Related Works}

\paragraph{3D grounding systems.}
Priors works have proposed methods for 3D grounding, leveraging context from the full 3D scene \cite{huang2021text, yuan2021instancerefer, zhao20213dvg, jain2022bottom}. Many of such methods take on an end-to-end approach, and jointly attend over language and 3D point clouds \cite{cai20223djcg, luo20223d, chen2022ham, chen2021d3net, chen2020scanrefer}. More specifically, several works leverage the Transformer~\cite{vaswani2017attention} architecture to solve the 3D referring expression comprehension task \cite{dailan2018transref, junha2021lanref, huang2022multi, abdelreheem20223dreftransformer, yang2021sat}. Notably, TransRefer3D \cite{dailan2018transref} uses a Transformer-based network to extract entity-and-relation-aware representations, and LanguageRefer \cite{junha2021lanref} employs a Transformer architecture over bounding box embeddings and language embedding from DistilBert \cite{sanh2019distilbert}. The Multi-View Transformer \cite{huang2022multi} projects the 3D scene to a multi-view space to learn robust representations. Other methods have used 2D information to augment grounding in 3D \cite{yang2021sat}, including LAR \cite{bakr2022look}, which uses 2D signals generated from 3D point clouds to assist its 3D encoder. Recently, NS3D~\cite{hsu2023ns3d} proposed a modular, neuro-symbolic approach to 3D grounding, which shows strong data efficiency and generalization compared to end-to-end works, but requires dense supervision.

All of the prior works for 3D grounding leverage classification labels in the training stage, many in the form of direct classification losses \cite{huang2022multi, hsu2023ns3d, bakr2022look, dailan2018transref, achlioptas2020referit3d}. BUTD-DETR \cite{jain2022bottom} and SAT \cite{yang2021sat} use outputs from pre-trained detection networks, including predicted bounding boxes and their class labels, as additional inputs. LanguageRefer \cite{junha2021lanref} applies a semantic classifier to obtain the class label of each object, which are tokenized and used as input. \citet{zhao20213dvg} use pre-trained segmentation networks to filter candidates based on predicted categories from
segmentation networks. While \model follows NS3D's neuro-symbolic concept learning paradigm, \model works in the naturally supervised setting, and is neither trained with classification labels nor uses classification predictions from other pre-trained networks.  

\vspace{-0.3cm}
\paragraph{Neuro-symbolic concept learners.}
Neuro-symbolic methods have shown strong visual reasoning performance and a wide range of capabilities. They parse visual reasoning queries into programs, and execute such programs with modular networks on a variety of visual domains \cite{andreas2016learning,johnson2017inferring,han2019visual,hudson2019learning,li2020closed, chen2021grounding, endo2023motion, hsu2023ns3d, hsu2023s, Mao2022PDSketch}. Neuro-Symbolic VQA~\cite{yi2018neural} first proposed program execution for 2D visual question answering, and the Neuro-Symbolic Concept Learner~\cite{NSCL} improved the training paradigm by removing direct supervision on scene representations and program traces in the 2D domain. A recent work, Logic-Enhanced Foundation Model~\cite{hsu2023s}, reduced prior knowledge required by replacing domain-specific languages with domain-independent logic. \model further lowers the guidance required for 3D concept learners by regularizing intermediate representations, enabling strong performance in naturally supervised settings with only 3D scene and QA pairs as supervision.

\vspace{-0.3cm}
\paragraph{Constraints in neural networks.}
Many works have proposed strategies for models to effectively encode knowledge-based rules, for the purposes of improving interpretability, sample efficiency, and compliance with safety constraints~\cite{von2021informed,deng2020integrating,giunchiglia2022deep,dash2022review}. \citet{von2021informed} describes existing methods which range in sources of constraints (\eg, expert knowledge), representations (\eg, graphs), and methods for integration (\eg, feature engineering). Most related to our work are methods that incorporate discrete rules into models' learning processes. Several works use rules as bases for model structures~\cite{li2019augmenting,marra2020relational,francca2014fast,towell1994knowledge}, while others learn latent embeddings of symbolic knowledge that can be naturally handled by neural networks~\cite{lensr,allamanis2017learning}. Logical rules have also been transformed into continuous and differentiable constraints, for example via t-norm \cite{diligenti2017integrating}, to serve as additional loss terms in training \cite{sl,sbr,stewart2017label,hoernle2022multiplexnet,hu2016harnessing}. Different from existing approaches, \model takes advantage of neuro-symbolic concept learners with structured representations, and utilizes language-based rules to regularize learned features. %
\section{Methods}
To describe \model, we first provide preliminary background on our task and method in Section~\ref{sec:background}. Then, we detail how \model distills constraints from LLMs in Section~\ref{sec:rule_gen}, and applies constraints on structured neuro-symbolic representations in Section~\ref{sec:losses}. After, we present our method's training details in Section~\ref{sec:training}. Finally, we discuss \model's language composition during inference in Section~\ref{sec:composition}.

\begin{figure*}[tp]
  \centering
    \includegraphics[width=1.0\linewidth]{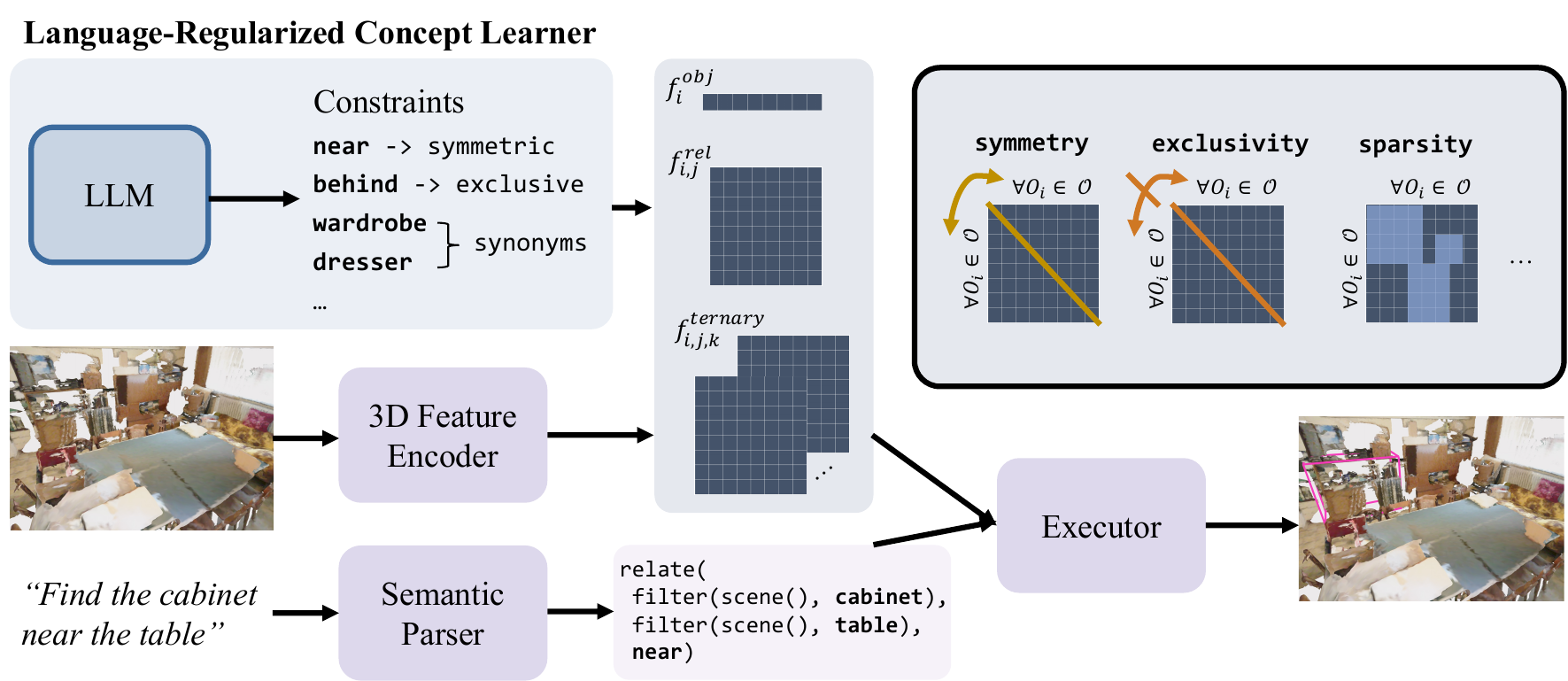}
  \caption{\model distills constraints from large language models, and injects these rules as regularization into the learning process of structured neuro-symbolic concept learners.}
\label{fig:systems}
\vspace{-0.5cm}
\end{figure*}

\subsection{Preliminaries}
\label{sec:background}
Neuro-symbolic concept learners \cite{NSCL, hsu2023ns3d, endo2023motion} are methods that decompose input language queries into symbolic programs, and differentiably execute the programs with concept grounding neural networks on the input visual modality. These approaches have shown strong 3D visual grounding capabilities, but often require dense supervision in complex domains, \eg, ground truth object bounding boxes and classification labels for objects in order to train intermediate programs. In this paper, we build \model from NS3D \cite{hsu2023ns3d}, a neuro-symbolic concept learner that conducts grounding in the 3D domain, in which such dense supervision is expensive and difficult to annotate. NS3D was proposed to tackle the task of 3D referring expression comprehension (3D-REC). 

\vspace{-0.3cm}
\paragraph{Problem statement.}
In 3D-REC, we are given a scene $\mathcal{S}$, which is represented as an RGB-colored point cloud of $C$ points $\mathcal{S} \in \mathbb{R}^{C \times 6}$, and an utterance $\mathcal{U}$ describing a target object and its relationship with other objects in $\mathcal{S}$. The goal is to localize the target object $\mathcal{T}$. Notably, there exists distractor objects of the same class category as $\mathcal{T}$, such that understanding the full referring expression is necessary in order to answer the query. For example, in a cluttered living room, $\mathcal{U}$ may be ``the chair beside the shelf'', which requires a neuro-symbolic concept learner to first find the \textit{shelf}, then find the object \textit{beside} it which is of the class \textit{chair}. There may be many \textit{chairs} in the room, from which the model must identify the target object from. %

\vspace{-0.3cm}
\paragraph{Concept learners.}
The overall architecture of \model follows that of NS3D, and its concept learning framework consists of three main components. The first is a semantic parser that parses the input language $\mathcal{U}$ into a symbolic program $\mathcal{P}$. The symbolic program consists of a hierarchy of primitive operations defined in a domain-specific language for 3D visual reasoning tasks, and represents the reasoning process underlying $\mathcal{U}$. Given the utterance example above, the semantic parser (here, a large language model) will yield the program \texttt{relate(filter(scene(), chair), filter(scene(), shelf), beside)}, which indicates the functions that should be run to output the answer.

The second is a 3D feature encoder that extracts structured object-centric features for each scene. The 3D-REC task gives segmented object point clouds for each scene as input, which NS3D leverages; however, \model instead uses VoteNet~\cite{qi2019deep} to reduce required annotations by detecting objects directly from $\mathcal{S}$. For each detected object point cloud of $M_i$ points $\mathcal{O}_{i} \in \mathbb{R}^{M_i \times 6}$, a 3D backbone (here, PointNet++ \cite{qi2017pointnet++}) takes $\mathcal{O}_{i}$ as input and outputs its corresponding feature $f^{\text{obj}}_{i} \in \mathbb{R}^d$, where $d$ is the dimension of the feature. For each pair and triple of objects, an encoder learns relational features $f^{\text{binary}}_{i,j}$ and $f^{\text{ternary}}_{i,j,k}$, which represent the binary relations (\eg, beside) and ternary relations (\eg, between) among objects respectively. Given $N$ objects in the scene, the unary representation $f^{\text{obj}}$ is a vector with $N$ features, one representing each object. The binary representation $f^{\text{binary}}$ is matrix of $N \times N$ features, representing relations between each pair of objects, and similarly for ternary features $f^{\text{ternary}}$. %

The third component of concept learners is a neural network-based program executor. The executor takes the program $\mathcal{P}$ and learned features $(f^{\text{obj}}, f^{\text{binary}}, f^{\text{ternary}})$, and returns the target object $\mathcal{T}$. During the execution of the symbolic program, the entity-centric features and learned concept embeddings will be used to compute score vectors, for example, $y^{\text{chair}}, y^{\text{shelf}} \in \mathbb{R}^N$; and a probability matrix, for example, $prob^{\text{beside}} \in \mathbb{R}^{N \times N}$, where $N$ denotes the number of objects in the scene. Elements of $y^{\text{chair}}$ denote the likelihood of objects' belonging to category \textit{chair}, and elements of $prob^{\text{beside}}$ denote the likelihood of object pairs' satisfying the relation \textit{beside}. Given the aforementioned symbolic program, the executor reasons with
$$
y = \text{min} \; (y^{\text{chair}}, prob^{\text{beside}} \times sx(y^{\text{shelf}})),
$$
where min is element-wise minumum and $sx()$ is the softmax function applied to $y^{\text{shelf}}$. Finally, the index of the referred object can be found with $\text{argmax} \; y$.

\vspace{-0.3cm}
\paragraph{Naturally supervised setting.}
In the canonical 3D-REC setup, class labels for each ground truth object bounding box in the scene are given, and used in all previous works as well as in NS3D to supervise intermediate programs. NS3D hence is trained with an object classification loss $\mathcal{L}_{cls}$ along with the final target object prediction loss $\mathcal{L}_{pred}$. In contrast, \model operates in a naturally supervised setting, supervised by only $\mathcal{L}_{pred}$ and a set of constraint-based regularization losses. In our low guidance setting, ground truth bounding boxes are not known, and we instead use VoteNet \cite{qi2019deep} to generate object detections. %

\vspace{-0.3cm}
\paragraph{Constraints.}
In order to learn in indirectly supervised settings, \model leverages language-based constraints as regularization. These constraints are based on language priors (\eg, derived from synonyms), hence distilled from LLMs, or are based on general priors (\eg, sparsity). We can enforce such rules on the structured and interpretable representations of concept learners, by way of regularization losses and data augmentation. We describe the definition and application of constraints below.

\subsection{Constraint Generation with LLMs}
\label{sec:rule_gen}
\model encodes language-based constraints from LLMs in concept learners. These constraints are generally applicable across all language-based tasks, regardless of input visual modality. To this end, \model uses LLMs to extract concepts names that satisfy a set of language-based rules. We propose symmetry~\cite{gleitman1996similar}, exclusivity~\cite{markman1988children}, and synonymity~\cite{miller1995wordnet} as general categories of language priors that capture a broad set of language properties and semantic contexts. At a high level, these rules specify the taxonomy of the concepts in relation to one another, as well as its effect on other concepts. Importantly, the constraints are broad, are applicable for all language queries, encode a wide range of properties, and can be used across different datasets. 

We query LLMs to classify concepts into these aforementioned constraints, by providing the set of concepts automatically extracted from the semantic parser, as well as definitions of the language rules. LLMs can accurately extract concepts that satisfy the constraints, as the models capture common usage of the concepts. Note that these rules can also be cheaply annotated by humans, compared to the dense supervision otherwise required in the form of 3D bounding boxes or classification labels. Below, we describe the definition for each category of language-based constraints, give examples of concepts from each category, and provide details on how we integrate the rule into \model.

\vspace{-0.3cm}
\paragraph{Symmetry.}
Relations between objects can be symmetric, in which the same relation holds when the order of the objects in the given relation is reversed. For example, given the relational concept \textit{close}, language priors dictate that the table \textit{close} to the chair implies that the chair is also \textit{close} to the table. Hence, the probability matrix $prob^{\text{close}}$ of size $N \times N$, which describes the likelihood of object pairs' satisfying the relation \textit{close}, should be symmetric. 

To distill a set of concepts that are symmetric, we prompt an LLM (here, GPT-3.5 \cite{brown2020language}) with the parsed set of relational concepts, and ask it to output the subset of concepts that exhibit such reciprocity. Given the definition of the constraint as prompt, the LLM is able to automatically and accurately determine whether the relational concept satisfies the symmetry prior. Examples of concepts extracted by the LLM include \textit{near}, \textit{beside}, \textit{far}, etc. Then, \model encodes the symmetric property with the proposed concepts as constraints to regularize \model's intermediate representations during training. We include prompts in the Appendix.

\vspace{-0.3cm}
\paragraph{Exclusivity.}
We define exclusive relations as concepts that indicate opposing relations when the order of objects is reversed. For example, given the concept \textit{above}, the box \textit{above} the cabinet implies that the cabinet \textit{cannot} be \textit{above} the box. By querying the LLM for relational concepts that are exclusive and enforcing such constraints during training, we encourage \model to learn relational representations that are consistent with language-based priors. For example, \model's $prob^{\text{above}}$ matrix should not yield high probabilities for the relationships ``box above cabinet'' and ``cabinet above box'' given the same scene. Examples of exclusive relations proposed by the LLM include \textit{left}, \textit{behind}, \textit{beneath}, etc. See Figure~\ref{fig:example} for visualizations of \model's learned concepts. 

\vspace{-0.3cm}
\paragraph{Synonymity.}
Humans have developed an extended set of vocabulary, in which there exist synonyms that represent visually similar 3D objects. End-to-end models typically leverage such nuanced taxonomy and word semantics through pre-trained language encoders, such as BERT \cite{devlin2018bert}. However, such integration of similar language properties is not explicitly modeled in neuro-symbolic frameworks, which ground modular symbols. To enable neuro-symbolic concept learners to encode these language priors in structured representations, we first query LLMs for visually similar synonyms within the concepts of object categories. For example, concepts such as \textit{wardrobe} and \textit{dresser}, as well as \textit{table} and \textit{dining table} are visually similar and synonymous. 

\model then encourages the object-centric representations for the similar concepts to be closer to one another. To do so, \model first parses utterances into symbolic programs, then selects for programs that contain concepts with an LLM-defined synonym. For each of these programs, we augment the original program with a synonymized version, with a randomly selected synonym concept generated by the LLM substituted in. This encourages the answer, and hence execution trace of \model, to be similar for synonyms.

\subsection{Constraints on structured representations} 
\label{sec:losses}

We introduce two core methods for incorporating constraints into structured neuro-symbolic representations. The first is through regularization losses, and the second through data augmentation. We describe both approaches below. Notably, \model's intermediate representations can be indexed by concept name and by arity, which enables effective injection of these constraints into the training process.

\vspace{-0.3cm}
\paragraph{Regularization losses.}
Recall that during the execution of neuro-symbolic programs, relations between objects are represented as probability matrices. Here, we use $prob^{\text{binary}} \in \mathbb{R}^{N \times N}$ and $prob^{\text{ternary}} \in \mathbb{R}^{N \times N \times N}$ to denote the probability matrices of binary and ternary relations respectively. Their elements are interpreted as the likelihood that the referred relation exists between the pair or triple of objects. For example, $prob^{\text{beside}}_{i,j}$ specifies the probability that object \textit{i} is beside object \textit{j}, and $prob^{\text{between}}_{i,j,k}$ specifies the probability that object \textit{i} is between objects \textit{j} and \textit{k}, where \textit{i, j, k} are indices of objects. Note that we mask diagonal elements, which represents objects' relations with itself. We ignore them not only in the calculation of losses, but also during execution. Based these notations, we introduce the following regularization losses.

We first define a constraint on the sparsity of the probability matrix $prob^{\text{rel}}$ for each relational concept. Due to the noise in VoteNet object detections, \model must learn to parse out bounding boxes that are not valid objects in the scene. Hence, we encourage $prob^{\text{rel}}$ to be sparse, keeping large values and ignoring small values. We treat small probability values, where the model is uncertain about the relation between objects, as noise in object bounding box predictions. Therefore, we inject a sparsity regularization loss to denoise \model's execution on the object-centric representations. The loss $\mathcal{L}_{spar}$ is applied to both binary and ternary relations, and encourages the probability matrices to be sparse as to remove noise from VoteNet object detections. The sparsity regularization loss is defined as
$$\mathcal{L}_{spar} = || prob^{\text{rel}} ||_{1}.$$

We then describe the symmetry regularization loss $\mathcal{L}_{sym}$, which encourages symmetric relations, as proposed by LLMs in the previous section, by enforcing symmetry on the relation matrix. At a high level, this regularization decreases the difference between $prob^{\text{rel}}_{i,j}$ and $prob^{\text{rel}}_{j,i}$, given that the order of the object does not affect the relation prediction. We define $\mathcal{L}_{sym}$ as
$$\mathcal{L}_{sym} = || prob^{\text{rel}} - (prob^{\text{rel}})^{T} ||^{2}_{2}.$$

Finally, we introduce the exclusivity regularization loss $\mathcal{L}_{excl}$. The exclusivity loss can be interpreted as the opposite of the symmetric loss, where $prob^{\text{rel}}_{i,j}$ and $prob^{\text{rel}}_{j,i}$ is encouraged to not both have high values for a given scene.  We define $\mathcal{L}_{excl}$ as
$$\mathcal{L}_{excl} = || \text{max} (0, prob^{\text{rel}}) \odot \text{max} (0, prob^{\text{rel}})^{T} ||_{1}.$$

Together, these constraint-based regularization losses are strong signals for neuro-symbolic concept learners in indirectly supervised settings.

\vspace{-0.3cm}
\paragraph{Data augmentation.}
\model also learns to encode objects that represent similar concepts to closer object-centric representations through data augmentation. \model augments symbolic programs by replacing parsed concepts with a randomly selected synonym concept, if exists. As an example, given the program \texttt{relate(filter(scene(), wardrobe), filter(scene(), cabinet), close)}, and similar concepts of \textit{wardrobe} and \textit{dresser} distilled from an LLM, we will augment the data with an additional program \texttt{relate(filter(scene(), dresser), filter(scene(), cabinet), close)}. This program is supervised with the same answer as the original program, given the same scene. The \model execution trace will hence be encouraged to be similar between programs with synonym concepts, leading to similar intermediate representations. %

\subsection{Training}
\label{sec:training}

We train \model in the naturally supervised 3D-REC setting, with only scene and question-answer pairs. \model is trained with the target object prediction loss $\mathcal{L}_{pred}$, a standard cross-entropy loss, along with the proposed regularization losses from the above sections, which do not require additional annotation. Overall, the loss can be computed as
$$\mathcal{L} =  \mathcal{L}_{pred} + \alpha \mathcal{L}_{sym} + \beta \mathcal{L}_{excl} + \gamma \mathcal{L}_{spar}.$$

\subsection{Language-based composition}
\label{sec:composition}
During inference, \model can also query LLMs for language rules when presented with novel concepts not seen during training. Given an unseen concept, for example \textit{center}, an LLM can specify how it can be composed from a set of learned concepts, automatically extracted by \model's semantic parser. As an example, the ternary relation of \textit{center} can be decomposed into a series of spatial \textit{left} and \textit{right} relations. Hence, \model can execute a program of ``couch in the \textit{center} of a lamp and a desk'' as a combination of: couch \textit{left} of lamp and \textit{right} of desk, or couch \textit{right} of lamp and \textit{left} of desk. Then, \model can use the learned probability matrices of $prob^{\text{left}}$ and $prob^{\text{right}}$ to compose $prob^{\text{center}}$ as 
$$prob^{\text{center}}_{i,j,k} = \text{max} \; ( prob^{\text{left}}_{i,j}+prob^{\text{right}}_{i,k}, prob^{\text{right}}_{i,j}+prob^{\text{left}}_{i,k} ).$$

Similarly, for an unseen concept combination of \textit{not} with a concept such as \textit{behind}, the LLM can specify execution with the learned \textit{front} concept. In practice, \model builds a lookup table of all antonym pairs queried from LLMs. When presented with new concepts, \model will query the lookup table and find the antonym concept to execute.

In our experiments, we choose a subset of relations that we know can be composed with learned concepts as a proof of concept. We can also manually specify such language rules to enable execution of new concepts. 

\section{Experiments}
We evaluate \model on the ReferIt3D benchmark \cite{achlioptas2020referit3d}, which tests 3D referring expression comprehension on the ScanNet dataset \cite{dai2017scannet}. We specifically focus on the SR3D setting that leverages spatially-oriented referential language, and measure accuracy by matching the predicted objects with the target objects. As we use VoteNet object detections, target objects are calculated as the VoteNet detection that has the highest IOU with the ground truth bounding box. Notably, we study the naturally supervised setting, where all models are not given object-level classification labels during training. Our indirectly supervised setting removes the need for dense annotations in the 3D domain.

In Section~\ref{sec:comp}, we compare \model to NS3D~\cite{hsu2023ns3d}, the prior state-of-the-art neuro-symbolic method. We show that simple constraints from language significantly improves the performance of NS3D, and retains all benefits that structured frameworks yield. We additionally present comparisons to end-to-end methods on a variety of metrics, and show qualitative visualizations of \model's learned concepts. In Section~\ref{sec:ablation}, we present ablations of \model's rules and train setting. More visualizations can be found in the Appendix.

\begin{figure*}[tp!]
  \centering
  \includegraphics[width=\linewidth]{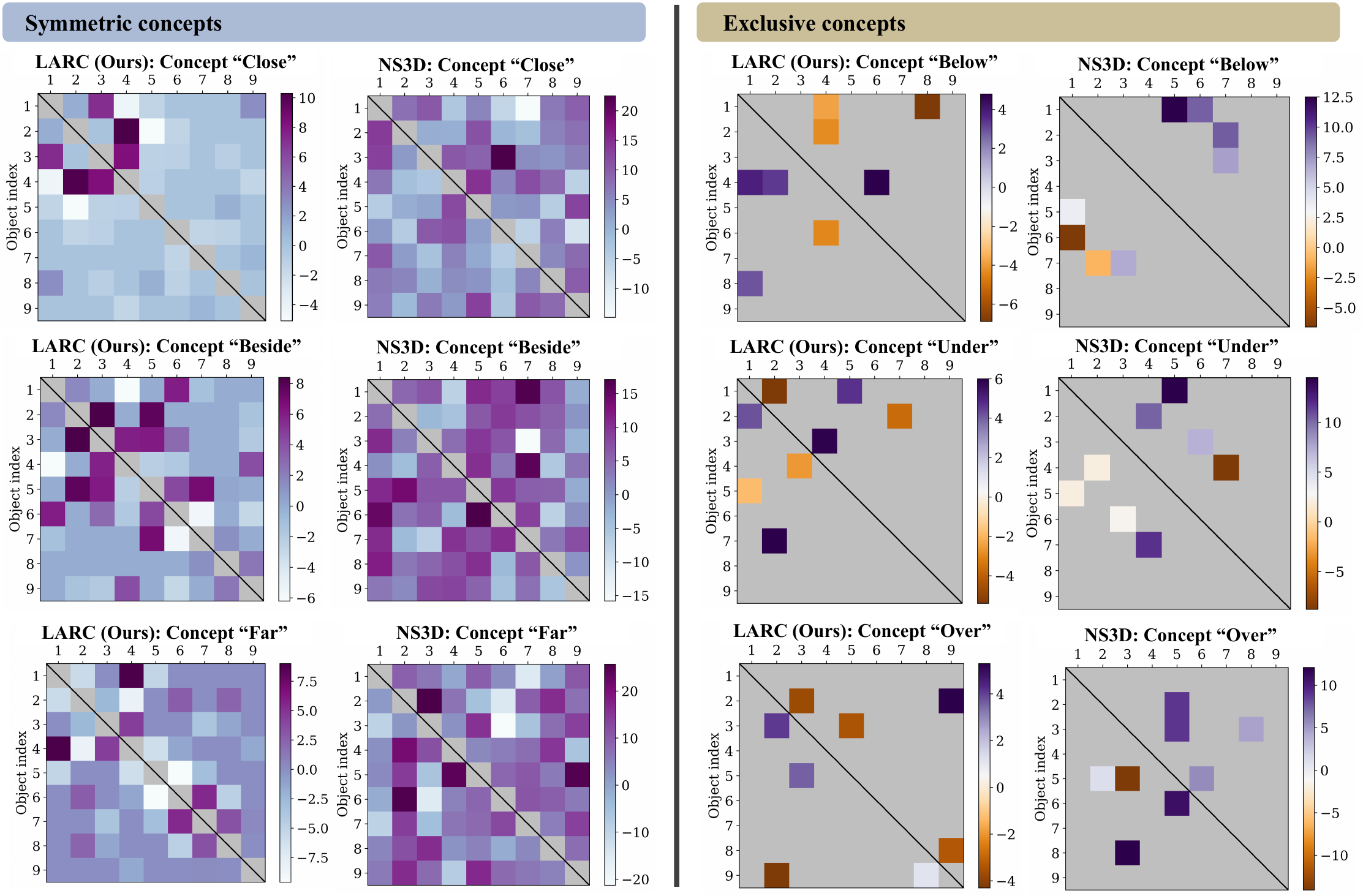}
  \caption{Visualizations of \model's and NS3D's learned features for symmetric (left two columns) and exclusive (right two columns) concepts; each matrix represents likelihood of pairs of objects' relations adhering to the given concept. \model features learn to encode constraints from language priors significantly more effectively than that of the NS3D baseline.}
\label{fig:example}
\vspace{-0.3cm}
\end{figure*}

\subsection{Comparison to prior work}
\label{sec:comp}
In this section, we evaluate \model and report test accuracy, generalization accuracy, data efficiency, and transfer accuracy on a new dataset. We compare \model with top-performing methods: BUTD-DETR \cite{jain2022bottom}, MVT \cite{huang2022multi}, NS3D \cite{hsu2023ns3d}, LAR \cite{bakr2022look}, TransRefer3D \cite{dailan2018transref}, and LanguageRefer \cite{junha2021lanref}. We additionally present qualitative visualizations of \model's learned concept representations in comparison to that of NS3D. We note that BUTD-DETR uses labels from pre-trained detectors directly as input to the model; we modified the architecture accordingly and
use our VoteNet detections.

\vspace{-0.3cm}
\paragraph{Accuracy.}
We first evaluate \model's performance on ReferIt3D, compared with prior methods. We also create two additional test subsets, which specifically report the accuracy of queries with symmetric and exclusive concepts. These subsets are selected from the original test set; the symmetric subset consists of $6{,}487$ examples, while the exclusive subset consists of $1{,}256$ examples.

In Table~\ref{table:accuracy}, we see that \model significantly outperforms the prior neuro-symbolic concept learner, NS3D, in the naturally supervised setting. 
\model improves performance of NS3D by $9$ point percent with our language regularization. \model also performs comparably to prior end-to-end methods, despite evaluation in the indirectly supervised setting, where neuro-symbolic concept learners tend to underperform. Importantly, we see a significant improvement in all other metrics of interest, described in sections below.

\setlength{\intextsep}{8pt}%
\setlength{\columnsep}{10pt}%

\begin{table}[h]
\small
  \centering
  \begin{tabular}{lllll}
    \toprule
         NS & Model & Test acc. & Sym. & Excl. \\
    \midrule
    \multirow{5}{*}{\xmark} 
    &LanguageRefer \cite{junha2021lanref} & $29.8$ & $29.1$ & $31.5$ \\
    &TransRefer3D \cite{dailan2018transref} & $30.6$ & $29.4$ & $32.9$ \\
    &LAR \cite{bakr2022look} & $32.2$ & $31.3$ & $34.5$ \\
    &MVT \cite{huang2022multi} & $35.7$ & $32.2$ & $37.3$  \\ 
    &BUTD-DETR \cite{jain2022bottom} & $\mathbf{38.5}$ & $\mathbf{36.3}$ & $\mathbf{40.2}$ \\
    \midrule
    \multirow{2}{*}{\cmark}&NS3D \cite{hsu2023ns3d} & $27.6$ & $24.9$ & $31.2$ \\
    &\model (Ours) & $\mathbf{36.6}$ & $\mathbf{34.5}$ & $\mathbf{38.4}$ \\
    \bottomrule
  \end{tabular}
  \caption{Comparison of \model with prior works in the naturally supervised setting of 3D referring expression comprehension. \model improves performance of NS3D significantly.}
  \label{table:accuracy}
\end{table}

\vspace{-0.3cm}
\paragraph{Generalization.}
We evaluate \model's ability to zero-shot generalize to unseen concepts based on language composition rules. We create two test sets with concepts not seen during training: ternary relations (\eg, \textit{center}, \textit{between}) and antonyms (\eg, \textit{not behind}, \textit{not left}). Queries with these concepts are removed from the train set. The ternary relations test set consists of $1{,}145$ examples from the test set, and the antonym test set consists of $50$ annotated examples.

In Table~\ref{table:zeroshot}, we see that it is easy to compose learned concepts to execute novel concepts in \model. In comparison, prior works suffer significantly and fail to generalize, even when leveraging word embeddings from powerful, pre-trained language encoders as input to the model.

\begin{table}[t]
\small
 \centering
  \begin{tabular}{lll}
    \toprule
         & Ternary acc. & Antonym acc. \\
    \midrule
    LanguageRefer \cite{junha2021lanref} & $8.9$ & $17.9$ \\
    TransRefer3D \cite{dailan2018transref} & $8.5$ & $18.2$ \\
    LAR \cite{bakr2022look} & $9.7$ & $21.8$ \\
    MVT \cite{huang2022multi} & $10.5$ & $22.0$ \\ 
    BUTD-DETR \cite{jain2022bottom} & $9.3$ & $19.5$ \\
    NS3D \cite{hsu2023ns3d} & $19.1$ & $16.0$ \\
    \model (Ours) & $\mathbf{37.4}$ & $\mathbf{32.6}$ \\
    \bottomrule
  \end{tabular}
  \caption{\model can use language-based rules to generalize execution of new concepts with composition of learned concepts.}
  \label{table:zeroshot}
\vspace{-0.5cm}
\end{table} 

\vspace{-0.3cm}
\paragraph{Data efficiency.}
We additionally demonstrate that \model retains strong data efficiency due to its modular concept learning framework. In Table~\ref{table:data-efficiency}, we see that \model is significantly more data efficient than prior works at $5\%$ ($1{,}423$ examples), $10\%$ ($2{,}846$ examples), $15\%$ ($4{,}268$ examples), $20\%$ ($5{,}691$ examples), and $25\%$ ($7{,}113$ examples) of train data used. Notably, \model sees a $6.8$ point percent gain from the top-performing prior work with $10\%$ of data.

\begin{table}[t]
\small
  \centering
  \begin{tabular}{llllll}
    \toprule
         & $5\%$ & $10\%$ & $15\%$ & $20\%$ & $25\%$ \\
    \midrule
    LanguageRefer \cite{junha2021lanref} & $17.7$ & $18.5$ & $19.7$ & $21.4$ & $22.8$ \\
    TransRefer3D \cite{dailan2018transref} & $16.4$ & $18.1$ & $19.5$ & $22.1$ & $23.3$ \\ 
    LAR \cite{bakr2022look} & $17.5$ & $19.9$ & $21.8$ & $22.5$ & $24.3$ \\
    MVT \cite{huang2022multi} & $19.5$ & $22.1$ & $24.2$ & $25.0$ & $27.0$ \\ 
    BUTD-DETR \cite{jain2022bottom} & $19.4$ & $22.3$ & $25.9$ & $28.2$ & $31.5$ \\
    NS3D \cite{hsu2023ns3d} & $15.7$ & $21.6$ & $22.2$ & $22.7$ & $23.1$ \\ 
    \model (Ours) & $\mathbf{24.9}$ & $\mathbf{29.1}$ & $\mathbf{32.4}$ & $\mathbf{33.9}$ & $\mathbf{34.8}$ \\
    \bottomrule
  \end{tabular}
  \caption{\model yields stronger data efficiency compared to prior works at five different percentage points of train data.}
  \label{table:data-efficiency}
\vspace{-0.5cm}
\end{table}
 
\vspace{-0.3cm}
\paragraph{Transfer to an unseen dataset.}
We also evaluate \model's transfer performance to an unseen dataset, ScanRefer \cite{chen2020scanrefer}, which contains new utterances on scenes from ScanNet \cite{dai2017scannet}. We retrieve a subset of $384$ ScanRefer examples that reference the same object categories and relations as in ReferIt3D, such that all methods can be run inference-only. %

In Table~\ref{table:scanrefer}, we present results to show that \model's neuro-symbolic framework enables effective transfer to the ScanRefer dataset, while only being trained on the ReferIt3D dataset. Prior methods, even the best end-to-end ones of BUTD-DETR \cite{jain2022bottom} and MVT \cite{huang2022multi}, do not enable such generalization. We see that \model outperforms BUTD-DETR by $14.8$ point percent and MVT by $15.2$ point percent.

\vspace{-0.3cm}
\paragraph{Qualitative visualizations.}
Our proposed regularization method enables \model to learn representations that are consistent with constraints from language properties. To examine this qualitatively, we present visualizations of \model's learned concepts in Figure~\ref{fig:example}. We visualize the probability matrix for each concept, where each value in the $N \times N$ matrix, $prob^{\text{rel}}_{i,j}$, represents the likelihood that the relation between objects of index $i$ and index $j$ adheres to the given concept. For exclusive concepts, we highlight high percentile values as well as their symmetric complements.

\begin{wraptable}{r}{0.25\textwidth}
\small
  \centering
  \begin{tabular}{ll}
    \toprule
         & ScanRefer \\
    \midrule
    LangRefer \cite{junha2021lanref} & $13.9$ \\
    TransRefer \cite{dailan2018transref} & $14.7$ \\
    LAR \cite{bakr2022look} & $15.4$ \\
    MVT \cite{huang2022multi} & $17.7$ \\ 
    BUTD \cite{jain2022bottom} & $18.1$ \\
    NS3D \cite{hsu2023ns3d} & $22.4$ \\
    \model (Ours) & $\mathbf{32.9}$ \\
    \bottomrule
  \end{tabular}
  \caption{\model enables transfer of learned concepts to the ScanRefer dataset \cite{chen2020scanrefer}, while prior works notably drop in performance.}
  \label{table:scanrefer}
\end{wraptable}

\iffalse
%
\begin{table}[ht]
\small
  \centering
  \begin{tabular}{ll}
    \toprule
         & ScanRefer accuracy \\
    \midrule
    MVT \cite{huang2022multi} & $17.7$ \\ 
    LAR \cite{bakr2022look} & $15.4$ \\
    TransRefer3D \cite{dailan2018transref} & $14.7$ \\ 
    LanguageRefer \cite{junha2021lanref} & $13.9$ \\
    NS3D \cite{hsu2023ns3d} & $22.4$ \\
    \model & $\mathbf{32.9}$ \\
    \bottomrule
  \end{tabular}
  \caption{\model enables transfer of learned concepts to the ScanRefer dataset \cite{chen2020scanrefer}, while prior works notably drop in performance.}
  \label{table:scanrefer}
\end{table}
%
\fi 

In the left two columns of Figure~\ref{fig:example}, we see that \model learns symmetric matrices for concepts in which object order in the relation does not matter, such as \textit{close} and \textit{far}. In contrast, NS3D's matrices are noisy and do not exhibit the same consistency. For concepts in which relations are reversed when objects are reversed in order, such as \textit{under} and \textit{over}, \model captures the exclusive relation between $prob^{\text{rel}}_{i,j}$ and $prob^{\text{rel}}_{j,i}$. In the right two columns of Figure~\ref{fig:example}, \model's matrices yield opposing values, and hence colors, in symmetric indices, while NS3D does not encode this knowledge in its representations.

\subsection{Ablations}
\label{sec:ablation}
Finally, we present ablations of \model's performance without each constraint. We additionally ablate \model's improvement on NS3D in settings with classification supervision.

\vspace{-0.3cm}
\paragraph{Constraints.}
In Table~\ref{table:loss}, we compare \model with different variants of \model trained without each constraint. We see that each of the general rules is important to encode in \model, as the removal of any constraint leads to worse performance. The synonym prior yields a strong effect on \model, while the sparsity prior affects \model at a smaller margin. We hypothesize that this is because the synonym prior is applied on concepts that encode object categories, which are more difficult to learn without classification supervision. Noise in VoteNet object detections may be more trivial in comparison.

\begin{wraptable}{r}{0.2\textwidth}
\small
  \centering
  \begin{tabular}{ll}
    \toprule
         & Acc. \\
    \midrule
    \model (Ours) & $36.6$ \\
    \midrule
    w/o Symmetry & $34.9$ \\
    w/o Exclusivity & $35.1$ \\ 
    w/o Sparsity & $35.8$ \\
    w/o Synonymity & $33.4$ \\
    \bottomrule
  \end{tabular}
  \caption{\model's accuracy without each constraint.}
  \label{table:loss}
\end{wraptable} 
\vspace{-0.3cm}
\paragraph{Supervision.}
In Table~\ref{table:box}, we present results on a train setting that gives models access to object-level classification labels, to evaluate \model's performance under denser supervision. The classification label of each VoteNet detected object is assigned the label of the ground truth object with the highest IOU to it. While \model still improves NS3D by $6.6$ point percent in this setting, we see less of a performance gain from our proposed regularization, compared to the $9$ point percent improvement in the naturally supervised setting. We hypothesize that this is because object-level classification supervision reduces uncertainty in \model's object-centric representations during training, hence lessens the need for regularization. We note that as the class labels for supervision are applied on predicted bounding boxes with potentially incomplete object point clouds, classification supervision does not yield significant improvements. 

\begin{table}[h]
\small
  \centering
  \begin{tabular}{lll|ll}
    \toprule
        & Cls. & Acc. &  Cls. & Acc. \\
    \midrule
    NS3D \cite{hsu2023ns3d}  & \cmark & $31.6$ & \xmark & $27.6$ \\
    \model (Ours) & \cmark & $\mathbf{38.2}$  & \xmark & $\mathbf{36.6}$ \\
    \bottomrule
  \end{tabular}
  \caption{Comparisons under a train setting with object-level classification supervision on predicted bounding boxes.}
  \label{table:box}
\vspace{-0.4cm}
\end{table}

%
\section{Conclusion}
3D visual grounding models perform poorly in naturally supervised settings, without access to object-level semantic labels or ground truth bounding boxes. We propose the Language-Regularized Concept Learner as a neuro-symbolic method that uses language regularization to significantly improve performance in indirectly supervised settings. We demonstrate that simple constraints, with concepts distilled from large language models, can significantly increase accuracy by way of regularization on structured representations. We show that \model is extremely generalizable, highly data efficient, and effective transfers learned concepts to different datasets, by only looking at 3D scene and question-answer pairs in the naturally supervised setting.

\vspace{-0.2cm}
\paragraph{Acknowledgments.}
This work is in part supported by NSF RI \#2211258, ONR N00014-23-1-2355, ONR YIP N00014-24-1-2117, and AFOSR YIP FA9550-23-1-0127.

\clearpage
{
    \small
    \bibliographystyle{ieeenat_fullname}
    \bibliography{main}

\begin{thebibliography}{55}
\providecommand{\natexlab}[1]{#1}
\providecommand{\url}[1]{\texttt{#1}}
\expandafter\ifx\csname urlstyle\endcsname\relax
  \providecommand{\doi}[1]{doi: #1}\else
  \providecommand{\doi}{doi: \begingroup \urlstyle{rm}\Url}\fi

\bibitem[Abdelreheem et~al.(2022)Abdelreheem, Upadhyay, Skorokhodov, Al~Yahya,
  Chen, and Elhoseiny]{abdelreheem20223dreftransformer}
Ahmed Abdelreheem, Ujjwal Upadhyay, Ivan Skorokhodov, Rawan Al~Yahya, Jun Chen,
  and Mohamed Elhoseiny.
\newblock {3DRefTransformer: Fine-grained Object Identification in Real-world
  Scenes Using Natural Language}.
\newblock In \emph{WACV}, pages 3941--3950, 2022.

\bibitem[Achlioptas et~al.(2020)Achlioptas, Abdelreheem, Xia, Elhoseiny, and
  Guibas]{achlioptas2020referit3d}
Panos Achlioptas, Ahmed Abdelreheem, Fei Xia, Mohamed Elhoseiny, and Leonidas
  Guibas.
\newblock {ReferIt3D: Neural Listeners for Fine-grained 3D Object
  Identification in Real-world Scenes}.
\newblock In \emph{ECCV}, pages 422--440. Springer, 2020.

\bibitem[Allamanis et~al.(2017)Allamanis, Chanthirasegaran, Kohli, and
  Sutton]{allamanis2017learning}
Miltiadis Allamanis, Pankajan Chanthirasegaran, Pushmeet Kohli, and Charles
  Sutton.
\newblock Learning continuous semantic representations of symbolic expressions.
\newblock In \emph{ICML}, 2017.

\bibitem[Andreas et~al.(2016)Andreas, Rohrbach, Darrell, and
  Klein]{andreas2016learning}
Jacob Andreas, Marcus Rohrbach, Trevor Darrell, and Dan Klein.
\newblock {Learning to Compose Neural Networks for Question Answering}.
\newblock In \emph{NAACL-HLT}, 2016.

\bibitem[Bakr et~al.(2022)Bakr, Alsaedy, and Elhoseiny]{bakr2022look}
Eslam Bakr, Yasmeen Alsaedy, and Mohamed Elhoseiny.
\newblock {Look Around and Refer: 2D Synthetic Semantics Knowledge Distillation
  for 3D Visual Grounding}.
\newblock In \emph{NeurIPS}, pages 37146--37158, 2022.

\bibitem[Brown et~al.(2020)Brown, Mann, Ryder, Subbiah, Kaplan, Dhariwal,
  Neelakantan, Shyam, Sastry, Askell, et~al.]{brown2020language}
Tom Brown, Benjamin Mann, Nick Ryder, Melanie Subbiah, Jared~D Kaplan, Prafulla
  Dhariwal, Arvind Neelakantan, Pranav Shyam, Girish Sastry, Amanda Askell,
  et~al.
\newblock {Language Models are Few-Shot Learners}.
\newblock \emph{NeurIPS}, 33:\penalty0 1877--1901, 2020.

\bibitem[Cai et~al.(2022)Cai, Zhao, Zhang, Sheng, and Xu]{cai20223djcg}
Daigang Cai, Lichen Zhao, Jing Zhang, Lu Sheng, and Dong Xu.
\newblock {3DJCG: A Unified Framework for Joint Dense Captioning and Visual
  Grounding on 3D Point Clouds}.
\newblock In \emph{CVPR}, pages 16464--16473, 2022.

\bibitem[Chen et~al.(2020)Chen, Chang, and Nie{\ss}ner]{chen2020scanrefer}
Dave~Zhenyu Chen, Angel~X Chang, and Matthias Nie{\ss}ner.
\newblock {ScanRefer: 3D Object Localization in RGB-D Scans using Natural
  Language}.
\newblock In \emph{ECCV}, pages 202--221. Springer, 2020.

\bibitem[Chen et~al.(2021{\natexlab{a}})Chen, Wu, Nießner, and
  Chang]{chen2021d3net}
Dave~Zhenyu Chen, Qirui Wu, Matthias Nießner, and Angel~X. Chang.
\newblock {D3Net: A Speaker-listener Architecture for Semi-supervised Dense
  Captioning and Visual Grounding in RGB-D Scans}, 2021{\natexlab{a}}.

\bibitem[Chen et~al.(2022)Chen, Luo, Wei, Ma, and Zhang]{chen2022ham}
Jiaming Chen, Weixin Luo, Xiaolin Wei, Lin Ma, and Wei Zhang.
\newblock {HAM: Hierarchical Attention Model with High Performance for 3D
  Visual Grounding}.
\newblock \emph{arXiv preprint arXiv:2210.12513}, 2022.

\bibitem[Chen et~al.(2021{\natexlab{b}})Chen, Mao, Wu, Wong, Tenenbaum, and
  Gan]{chen2021grounding}
Zhenfang Chen, Jiayuan Mao, Jiajun Wu, Kwan-Yee~Kenneth Wong, Joshua~B
  Tenenbaum, and Chuang Gan.
\newblock {Grounding Physical Concepts of Objects and Events Through Dynamic
  Visual Reasoning}.
\newblock In \emph{ICLR}, 2021{\natexlab{b}}.

\bibitem[Dai et~al.(2017)Dai, Chang, Savva, Halber, Funkhouser, and
  Nie{\ss}ner]{dai2017scannet}
Angela Dai, Angel~X Chang, Manolis Savva, Maciej Halber, Thomas Funkhouser, and
  Matthias Nie{\ss}ner.
\newblock {ScanNet: Richly-Annotated 3D Reconstructions of Indoor Scenes}.
\newblock In \emph{CVPR}, pages 5828--5839, 2017.

\bibitem[Dash et~al.(2022)Dash, Chitlangia, Ahuja, and
  Srinivasan]{dash2022review}
Tirtharaj Dash, Sharad Chitlangia, Aditya Ahuja, and Ashwin Srinivasan.
\newblock A review of some techniques for inclusion of domain-knowledge into
  deep neural networks.
\newblock \emph{Scientific Reports}, 12\penalty0 (1):\penalty0 1040, 2022.

\bibitem[Deng et~al.(2020)Deng, Ji, Rainey, Zhang, and Lu]{deng2020integrating}
Changyu Deng, Xunbi Ji, Colton Rainey, Jianyu Zhang, and Wei Lu.
\newblock Integrating machine learning with human knowledge.
\newblock \emph{Iscience}, 23\penalty0 (11), 2020.

\bibitem[Devlin et~al.(2019)Devlin, Chang, Lee, and Toutanova]{devlin2018bert}
Jacob Devlin, Ming-Wei Chang, Kenton Lee, and Kristina Toutanova.
\newblock {BERT: Pre-training of Deep Bidirectional Transformers for Language
  Understanding}.
\newblock In \emph{ACL}, 2019.

\bibitem[Diligenti et~al.(2015)Diligenti, Gori, and Saccà]{sbr}
Michelangelo Diligenti, Marco Gori, and Claudio Saccà.
\newblock {Semantic-based Regularization for Learning and Inference}.
\newblock In \emph{Artificial Intelligence 244 (2017) 143–165}, 2015.

\bibitem[Diligenti et~al.(2017)Diligenti, Roychowdhury, and
  Gori]{diligenti2017integrating}
Michelangelo Diligenti, Soumali Roychowdhury, and Marco Gori.
\newblock Integrating prior knowledge into deep learning.
\newblock In \emph{International Conference on Machine Learning and
  Applications}, 2017.

\bibitem[Endo et~al.(2023)Endo, Hsu, Li, and Wu]{endo2023motion}
Mark Endo, Joy Hsu, Jiaman Li, and Jiajun Wu.
\newblock Motion question answering via modular motion programs.
\newblock \emph{ICML}, 2023.

\bibitem[Fran{\c{c}}a et~al.(2014)Fran{\c{c}}a, Zaverucha, and d’Avila
  Garcez]{francca2014fast}
Manoel~VM Fran{\c{c}}a, Gerson Zaverucha, and Artur~S d’Avila Garcez.
\newblock Fast relational learning using bottom clause propositionalization
  with artificial neural networks.
\newblock \emph{Machine learning}, 94:\penalty0 81--104, 2014.

\bibitem[Giunchiglia et~al.(2022)Giunchiglia, Stoian, and
  Lukasiewicz]{giunchiglia2022deep}
Eleonora Giunchiglia, Mihaela~Catalina Stoian, and Thomas Lukasiewicz.
\newblock Deep learning with logical constraints.
\newblock In \emph{IJCAI}, 2022.

\bibitem[Gleitman et~al.(1996)Gleitman, Gleitman, Miller, and
  Ostrin]{gleitman1996similar}
Lila~R Gleitman, Henry Gleitman, Carol Miller, and Ruth Ostrin.
\newblock Similar, and similar concepts.
\newblock \emph{Cognition}, 58\penalty0 (3):\penalty0 321--376, 1996.

\bibitem[Han et~al.(2019)Han, Mao, Gan, Tenenbaum, and Wu]{han2019visual}
Chi Han, Jiayuan Mao, Chuang Gan, Josh Tenenbaum, and Jiajun Wu.
\newblock {Visual Concept-Metaconcept Learning}.
\newblock In \emph{NeurIPS}, 2019.

\bibitem[He et~al.(2021)He, Zhao, Luo, Hui, Huang, Zhang, and
  Liu]{dailan2018transref}
Dailan He, Yusheng Zhao, Junyu Luo, Tianrui Hui, Shaofei Huang, Aixi Zhang, and
  Si Liu.
\newblock {TransRefer3D: Entity-and-Relation Aware Transformer for Fine-Grained
  3D Visual Grounding}.
\newblock In \emph{ACM International Conference on Multimedia}, pages
  2344--2352, 2021.

\bibitem[Hoernle et~al.(2022)Hoernle, Karampatsis, Belle, and
  Gal]{hoernle2022multiplexnet}
Nick Hoernle, Rafael~Michael Karampatsis, Vaishak Belle, and Kobi Gal.
\newblock {MultiplexNet: Towards Fully Satisfied Logical Constraints in Neural
  Networks}.
\newblock In \emph{AAAI}, pages 5700--5709, 2022.

\bibitem[Hsu et~al.(2023{\natexlab{a}})Hsu, Mao, Tenenbaum, and Wu]{hsu2023s}
Joy Hsu, Jiayuan Mao, Joshua~B Tenenbaum, and Jiajun Wu.
\newblock {What's Left? Concept Grounding with Logic-Enhanced Foundation
  Models}.
\newblock \emph{NeurIPS}, 2023{\natexlab{a}}.

\bibitem[Hsu et~al.(2023{\natexlab{b}})Hsu, Mao, and Wu]{hsu2023ns3d}
Joy Hsu, Jiayuan Mao, and Jiajun Wu.
\newblock {NS3D: Neuro-Symbolic Grounding of 3D Objects and Relations}.
\newblock In \emph{CVPR}, pages 2614--2623, 2023{\natexlab{b}}.

\bibitem[Hu et~al.(2016)Hu, Ma, Liu, Hovy, and Xing]{hu2016harnessing}
Zhiting Hu, Xuezhe Ma, Zhengzhong Liu, Eduard Hovy, and Eric Xing.
\newblock {Harnessing Deep Neural Networks with Logic Rules}.
\newblock In \emph{ACL}, 2016.

\bibitem[Huang et~al.(2021)Huang, Lee, Chen, and Liu]{huang2021text}
Pin-Hao Huang, Han-Hung Lee, Hwann-Tzong Chen, and Tyng-Luh Liu.
\newblock {Text-guided Graph Neural Networks for Referring 3D Instance
  Segmentation}.
\newblock In \emph{AAAI}, pages 1610--1618, 2021.

\bibitem[Huang et~al.(2022)Huang, Chen, Jia, and Wang]{huang2022multi}
Shijia Huang, Yilun Chen, Jiaya Jia, and Liwei Wang.
\newblock {Multi-View Transformer for 3D Visual Grounding}.
\newblock In \emph{CVPR}, 2022.

\bibitem[Hudson and Manning(2019)]{hudson2019learning}
Drew Hudson and Christopher~D Manning.
\newblock {Learning by Abstraction: The Neural State Machine}.
\newblock In \emph{NeurIPS}, 2019.

\bibitem[Jain et~al.(2022)Jain, Gkanatsios, Mediratta, and
  Fragkiadaki]{jain2022bottom}
Ayush Jain, Nikolaos Gkanatsios, Ishita Mediratta, and Katerina Fragkiadaki.
\newblock {Bottom Up Top Down Detection Transformers for Language Grounding in
  Images and Point Clouds}.
\newblock In \emph{ECCV}, pages 417--433. Springer, 2022.

\bibitem[Johnson et~al.(2017)Johnson, Hariharan, Van Der~Maaten, Hoffman,
  Fei-Fei, Lawrence~Zitnick, and Girshick]{johnson2017inferring}
Justin Johnson, Bharath Hariharan, Laurens Van Der~Maaten, Judy Hoffman, Li
  Fei-Fei, C Lawrence~Zitnick, and Ross Girshick.
\newblock {Inferring and Executing Programs for Visual Reasoning}.
\newblock In \emph{ICCV}, 2017.

\bibitem[Li et~al.(2020)Li, Huang, Hong, Chen, Wu, and Zhu]{li2020closed}
Qing Li, Siyuan Huang, Yining Hong, Yixin Chen, Ying~Nian Wu, and Song-Chun
  Zhu.
\newblock {Closed Loop Neural-Symbolic Learning via Integrating Neural
  Perception, Grammar Parsing, and Symbolic Reasoning}.
\newblock In \emph{ICML}, 2020.

\bibitem[Li and Srikumar(2019)]{li2019augmenting}
Tao Li and Vivek Srikumar.
\newblock {Augmenting Neural Networks with First-Order Logic}.
\newblock \emph{ACL}, 2019.

\bibitem[Luo et~al.(2022)Luo, Fu, Kong, Gao, Ren, Shen, Xia, and
  Liu]{luo20223d}
Junyu Luo, Jiahui Fu, Xianghao Kong, Chen Gao, Haibing Ren, Hao Shen, Huaxia
  Xia, and Si Liu.
\newblock {3D-SPS: Single-stage 3D Visual Grounding via Referred Point
  Progressive Selection}.
\newblock In \emph{CVPR}, pages 16454--16463, 2022.

\bibitem[Mao et~al.(2019)Mao, Gan, Kohli, Tenenbaum, and Wu]{NSCL}
Jiayuan Mao, Chuang Gan, Pushmeet Kohli, Joshua~B Tenenbaum, and Jiajun Wu.
\newblock {The Neuro-Symbolic Concept Learner: Interpreting Scenes, Words, and
  Sentences From Natural Supervision}.
\newblock In \emph{ICLR}, 2019.

\bibitem[Mao et~al.(2022)Mao, Lozano-Perez, Tenenbaum, and
  Kaelbing]{Mao2022PDSketch}
Jiayuan Mao, Tomas Lozano-Perez, Joshua~B. Tenenbaum, and Leslie~Pack Kaelbing.
\newblock {PDSketch: Integrated Domain Programming, Learning, and Planning}.
\newblock In \emph{NeurIPS}, 2022.

\bibitem[Markman and Wachtel(1988)]{markman1988children}
Ellen~M Markman and Gwyn~F Wachtel.
\newblock Children's use of mutual exclusivity to constrain the meanings of
  words.
\newblock \emph{Cognitive psychology}, 20\penalty0 (2):\penalty0 121--157,
  1988.

\bibitem[Marra et~al.(2020)Marra, Diligenti, Giannini, Gori, and
  Maggini]{marra2020relational}
Giuseppe Marra, Michelangelo Diligenti, Francesco Giannini, Marco Gori, and
  Marco Maggini.
\newblock Relational neural machines.
\newblock In \emph{ECAI}, 2020.

\bibitem[Miller(1995)]{miller1995wordnet}
George~A Miller.
\newblock Wordnet: a lexical database for english.
\newblock \emph{Communications of the ACM}, 38\penalty0 (11):\penalty0 39--41,
  1995.

\bibitem[Qi et~al.(2017)Qi, Yi, Su, and Guibas]{qi2017pointnet++}
Charles~Ruizhongtai Qi, Li Yi, Hao Su, and Leonidas~J Guibas.
\newblock {PointNet++: Deep Hierarchical Feature Learning on Point Sets in a
  Metric Space}.
\newblock \emph{NeurIPS}, 30, 2017.

\bibitem[Qi et~al.(2019)Qi, Litany, He, and Guibas]{qi2019deep}
Charles~R Qi, Or Litany, Kaiming He, and Leonidas~J Guibas.
\newblock {Deep Hough Voting for 3D Object Detection in Point Clouds}.
\newblock In \emph{CVPR}, pages 9277--9286, 2019.

\bibitem[Roh et~al.(2022)Roh, Desingh, Farhadi, and Fox]{junha2021lanref}
Junha Roh, Karthik Desingh, Ali Farhadi, and Dieter Fox.
\newblock {LanguageRefer: Spatial-Language Model for 3D Visual Grounding}.
\newblock In \emph{CoRL}, pages 1046--1056. PMLR, 2022.

\bibitem[Sanh et~al.(2019)Sanh, Debut, Chaumond, and Wolf]{sanh2019distilbert}
Victor Sanh, Lysandre Debut, Julien Chaumond, and Thomas Wolf.
\newblock {DistilBERT, a Distilled Version of BERT: Smaller, Faster, Cheaper
  and Lighter}.
\newblock \emph{arXiv preprint arXiv:1910.01108}, 2019.

\bibitem[Stewart and Ermon(2017)]{stewart2017label}
Russell Stewart and Stefano Ermon.
\newblock Label-free supervision of neural networks with physics and domain
  knowledge.
\newblock In \emph{Proceedings of the AAAI Conference on Artificial
  Intelligence}, 2017.

\bibitem[Towell and Shavlik(1994)]{towell1994knowledge}
Geoffrey~G Towell and Jude~W Shavlik.
\newblock Knowledge-based artificial neural networks.
\newblock \emph{Artificial intelligence}, 70\penalty0 (1-2):\penalty0 119--165,
  1994.

\bibitem[Vaswani et~al.(2017)Vaswani, Shazeer, Parmar, Uszkoreit, Jones, Gomez,
  Kaiser, and Polosukhin]{vaswani2017attention}
Ashish Vaswani, Noam Shazeer, Niki Parmar, Jakob Uszkoreit, Llion Jones,
  Aidan~N Gomez, {\L}ukasz Kaiser, and Illia Polosukhin.
\newblock {Attention is All You Need}.
\newblock \emph{NeurIPS}, 30, 2017.

\bibitem[Von~Rueden et~al.(2021)Von~Rueden, Mayer, Beckh, Georgiev,
  Giesselbach, Heese, Kirsch, Pfrommer, Pick, Ramamurthy,
  et~al.]{von2021informed}
Laura Von~Rueden, Sebastian Mayer, Katharina Beckh, Bogdan Georgiev, Sven
  Giesselbach, Raoul Heese, Birgit Kirsch, Julius Pfrommer, Annika Pick,
  Rajkumar Ramamurthy, et~al.
\newblock Informed machine learning--a taxonomy and survey of integrating prior
  knowledge into learning systems.
\newblock \emph{Transactions on Knowledge and Data Engineering}, 35\penalty0
  (1):\penalty0 614--633, 2021.

\bibitem[Wong et~al.(2023)Wong, Grand, Lew, Goodman, Mansinghka, Andreas, and
  Tenenbaum]{wong2023word}
Lionel Wong, Gabriel Grand, Alexander~K Lew, Noah~D Goodman, Vikash~K
  Mansinghka, Jacob Andreas, and Joshua~B Tenenbaum.
\newblock {From Word Models to World Models: Translating from Natural Language
  to the Probabilistic Language of Thought}.
\newblock \emph{arXiv preprint arXiv:2306.12672}, 2023.

\bibitem[Xie et~al.(2019)Xie, Xu, S.~Kankanhalli, S.~Meel, and Soh]{lensr}
Yaqi Xie, Ziwei Xu, Mohan S.~Kankanhalli, Kuldeep S.~Meel, and Harold Soh.
\newblock {Embedding Symbolic Knowledge into Deep Networks}.
\newblock In \emph{NeurIPS}, 2019.

\bibitem[Xu et~al.(2018)Xu, zhang, Friedman, Liang, and den Broeck]{sl}
Jingyi Xu, Zilu zhang, Tal Friedman, Yitao Liang, and Guy~Van den Broeck.
\newblock {A Semantic Loss Function for Deep Learning with Symbolic Knowledge}.
\newblock In \emph{ICML}, 2018.

\bibitem[Yang et~al.(2021)Yang, Zhang, Wang, and Luo]{yang2021sat}
Zhengyuan Yang, Songyang Zhang, Liwei Wang, and Jiebo Luo.
\newblock {SAT: 2D Semantics Assisted Training for 3D Visual Grounding}.
\newblock In \emph{ICCV}, pages 1856--1866, 2021.

\bibitem[Yi et~al.(2018)Yi, Wu, Gan, Torralba, Kohli, and
  Tenenbaum]{yi2018neural}
Kexin Yi, Jiajun Wu, Chuang Gan, Antonio Torralba, Pushmeet Kohli, and Josh
  Tenenbaum.
\newblock {Neural-Symbolic VQA: Disentangling Reasoning from Vision and
  Language Understanding}.
\newblock In \emph{NeurIPS}, 2018.

\bibitem[Yuan et~al.(2021)Yuan, Yan, Liao, Zhang, Wang, Li, and
  Cui]{yuan2021instancerefer}
Zhihao Yuan, Xu Yan, Yinghong Liao, Ruimao Zhang, Sheng Wang, Zhen Li, and
  Shuguang Cui.
\newblock {InstanceRefer: Cooperative Holistic Understanding for Visual
  Grounding on Point Clouds through Instance Multi-level Contextual Referring}.
\newblock In \emph{ICCV}, pages 1791--1800, 2021.

\bibitem[Zhao et~al.(2021)Zhao, Cai, Sheng, and Xu]{zhao20213dvg}
Lichen Zhao, Daigang Cai, Lu Sheng, and Dong Xu.
\newblock {3DVG-Transformer: Relation Modeling for Visual Grounding on Point
  Clouds}.
\newblock In \emph{ICCV}, pages 2928--2937, 2021.

\end{thebibliography}
}

\maketitlesupplementary

\appendix

The appendix is organized as the following. In Appendix~\ref{app:prompts}, we specify the prompts used to query LLMs in \model. In Appendix~\ref{app:viz}, we present additional visualizations of \model's performance. In Appendix~\ref{app:noise}, we include additional results of \model on different levels of box prediction noise.
In Appendix~\ref{app:scanrefer}, we discuss the ScanRefer~\cite{chen2020scanrefer} dataset.

\section{Prompts}
\label{app:prompts}

Below, we provide the prompts used to query large language models, specifically, GPT-3.5 \cite{brown2020language}, for concepts that satisfy \model's constraints. \\

\noindent \textbf{Symmetry and exclusivity}. \; We use the following prompt to categorize relational concepts, where \textit{[relations]} is the list of relational concepts automatically extracted from the input language by \model's semantic parser: 

\noindent \textit{We define two kinds of spatial relations: Asymmetric relations are relations that don't exhibit reciprocity when the order of the objects is reversed. Symmetric relations are relations that exhibit reciprocity when the order of the objects is reversed. Here are some relations: [relations]. For each relation, specify whether it is a symmetric relation or an asymmetric relation.} \\

\noindent \textbf{Synonyms}. \; We use the following two-round query to find visually similar synonyms in object categories, where the \textit{[object categories]} list is automatically extracted: 

\noindent First round: \textit{Here are some object categories: [object categories]. List categories that have similar meanings.} 

\noindent Second round: \textit{Within each group, list categories that have similar appearances.} \\

\section{Visualizations}
\label{app:viz}
In this section, we present additional visualizations of \model's performance. First, we compare \model's predictions to that of prior works on the ReferIt3D~\cite{achlioptas2020referit3d} dataset. Then, we provide execution trace examples of \model. After, we demonstrate failure cases of \model and include analyses. Finally, we show examples of VoteNet~\cite{qi2019deep} object detections in comparison to ground truth bounding boxes.

\begin{figure*}[tp]
  \centering
    \includegraphics[width=0.95\linewidth]{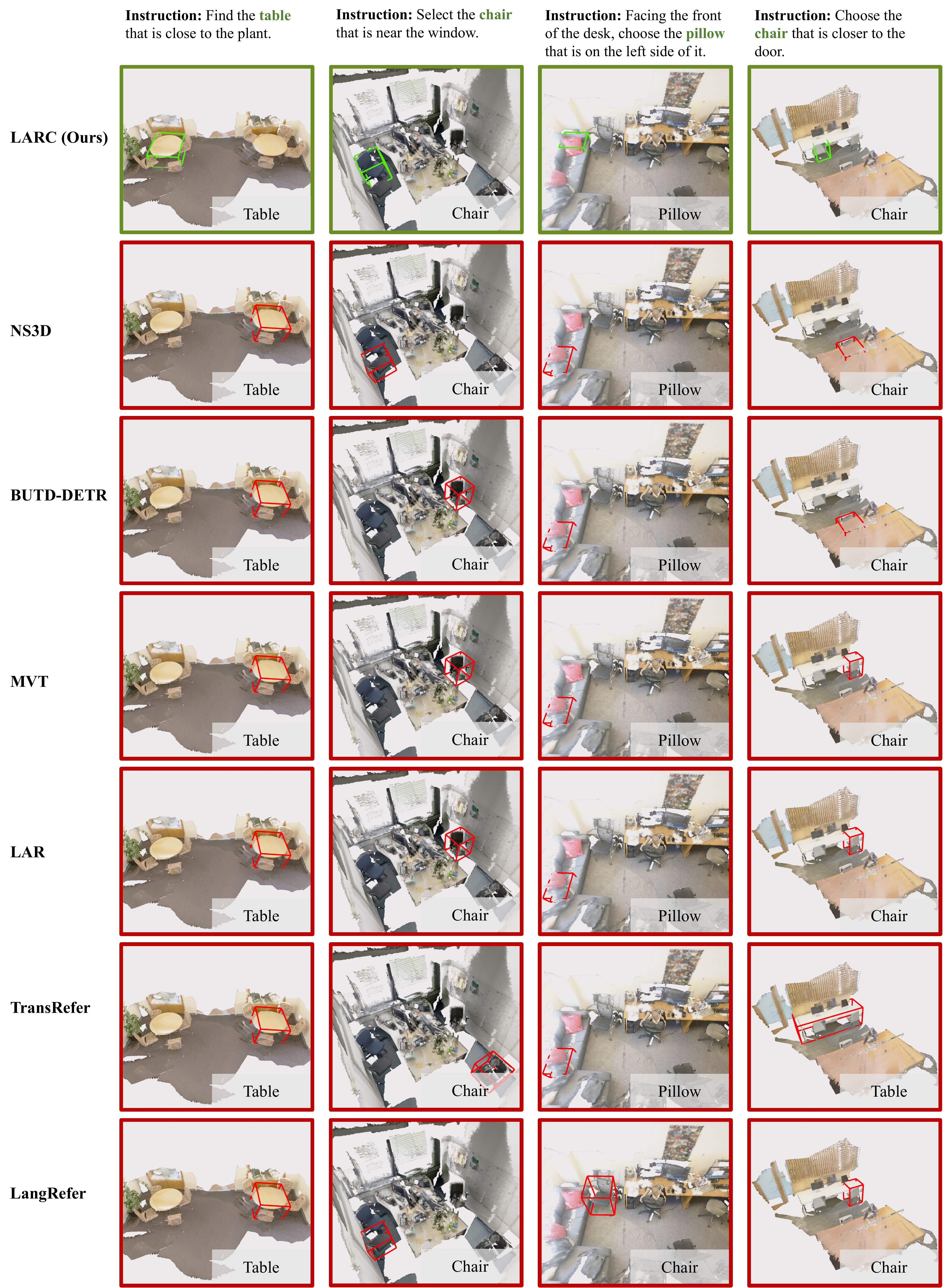}
  \caption{\model's performance compared to prior works in the naturally supervised setting; each column shows every model's prediction for a given instruction.}
\label{fig:app_comparison}
\end{figure*}

\paragraph{Comparison to prior works}
We present examples of \model's predictions as well as baselines' on the ReferIt3D~\cite{achlioptas2020referit3d} dataset. We see samples in Figure~\ref{fig:app_comparison} where \model outperforms baselines, including NS3D \cite{hsu2023ns3d}, BUTD-DETR \cite{jain2022bottom}, MVT \cite{huang2022multi}, LAR \cite{bakr2022look}, TransRefer \cite{dailan2018transref}, and LangRefer \cite{junha2021lanref}, in the naturally supervised 3D grounding setting.

\begin{figure*}[tp]
  \centering
    \includegraphics[width=1.0\linewidth]{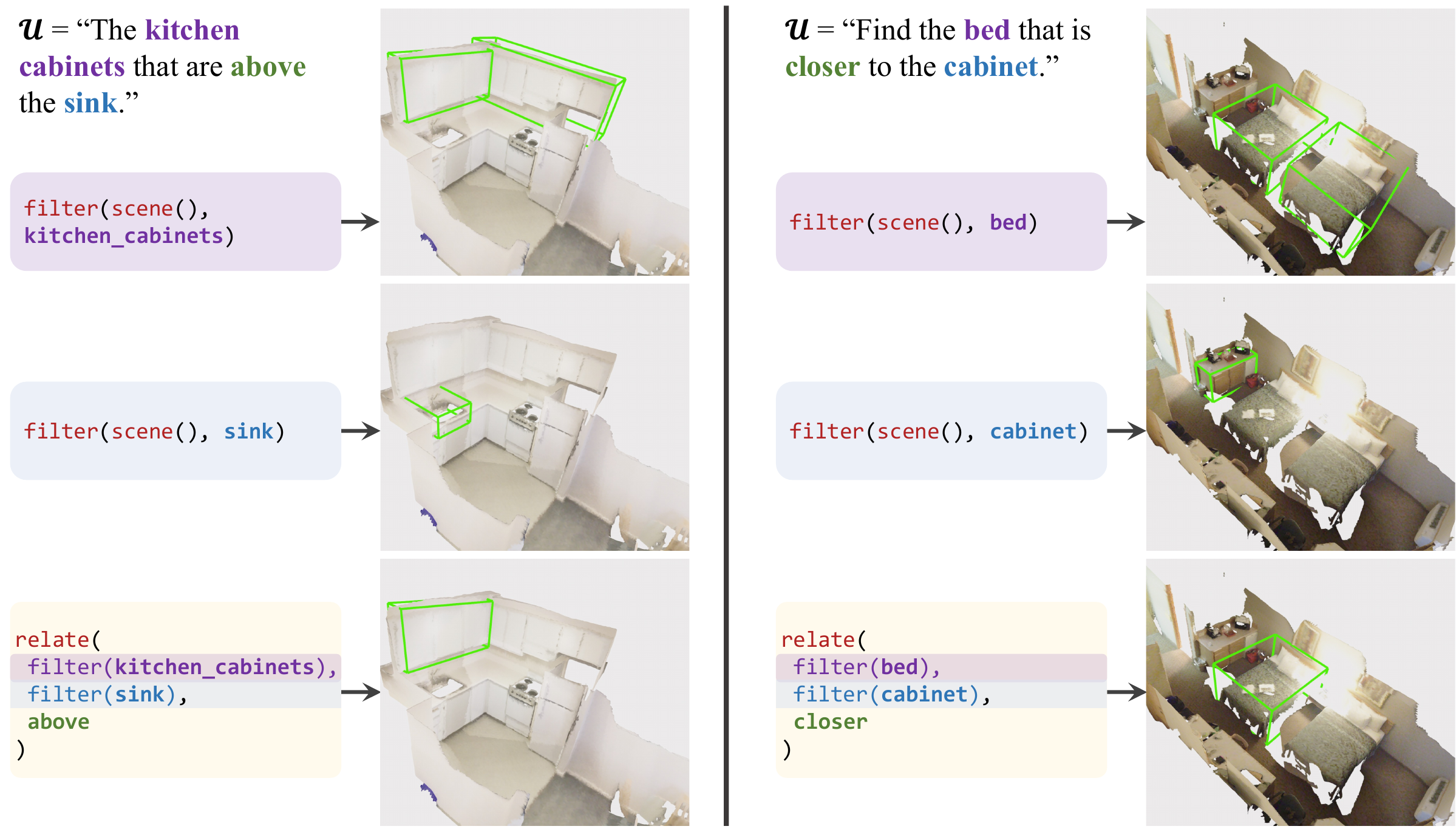}
  \caption{\model's neuro-symbolic framework executes symbolic programs hierarchically to retrieve the target answers.}
\label{fig:app_trace}
\end{figure*}

\paragraph{Execution traces}
In Figure~\ref{fig:app_trace}, we present examples of \model's execution trace. \model first parses input instruction utterances into symbolic programs, then hierarchically executes each modular program to retrieve the answer.

\begin{figure*}[tp]
  \centering
    \includegraphics[width=1.0\linewidth]{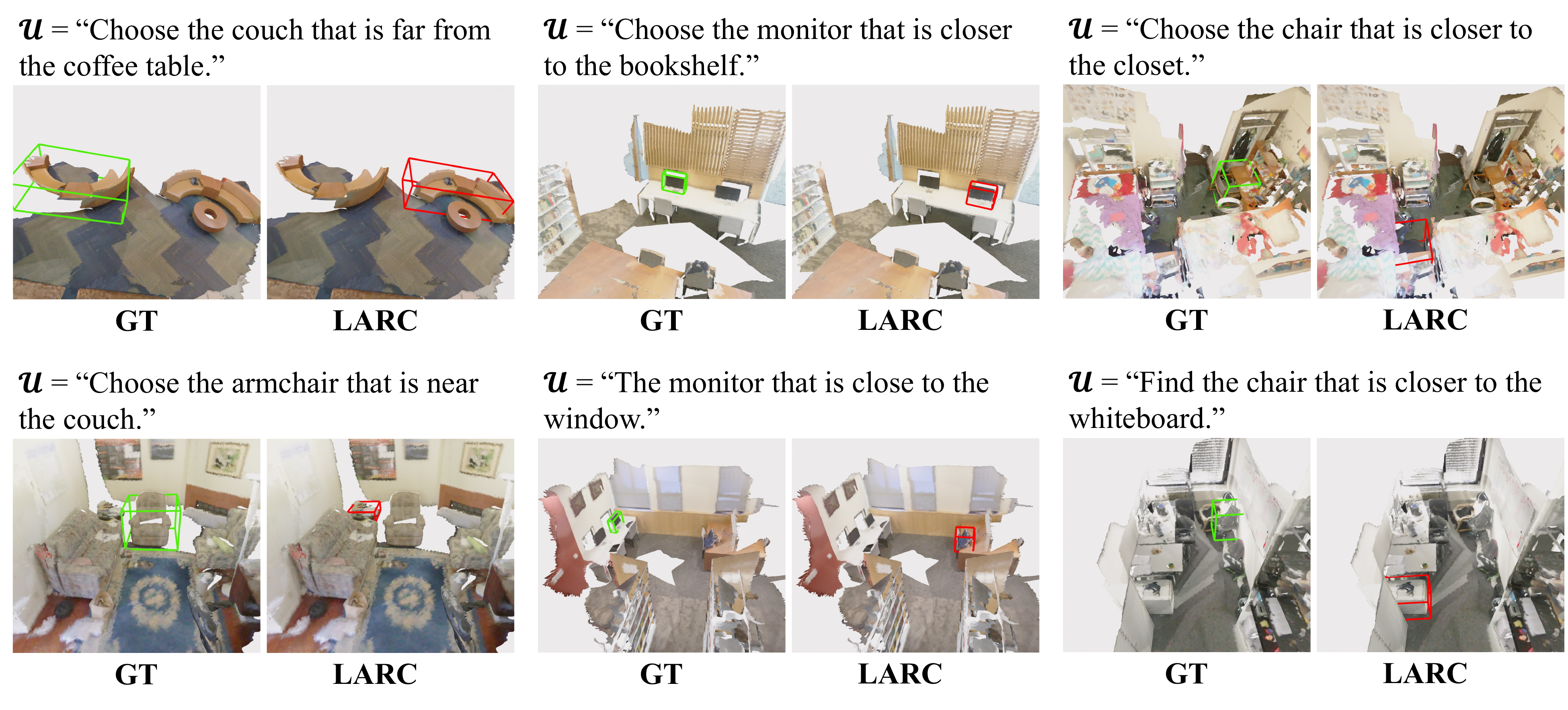}
  \caption{\model can fail in understanding 3D relations (top row) or 3D object categories (bottom row); its modularity enables such analyses.}
\label{fig:app_failures}
\end{figure*}

\paragraph{Failure cases}
We provide several examples of \model's failure cases in Figure~\ref{fig:app_failures}. In the top row, we see cases where \model finds target objects of the correct object category, but with incorrect relations. In the bottom row, we see cases where \model yields target objects of incorrect object categories. \model is likely to fail in 3D visual grounding when the target object category is one without data-augumented synonyms during training, as it is difficult to learn with few examples in the naturally supervised setting.

\begin{figure*}[t!]
  \centering
    \includegraphics[width=1.0\linewidth]{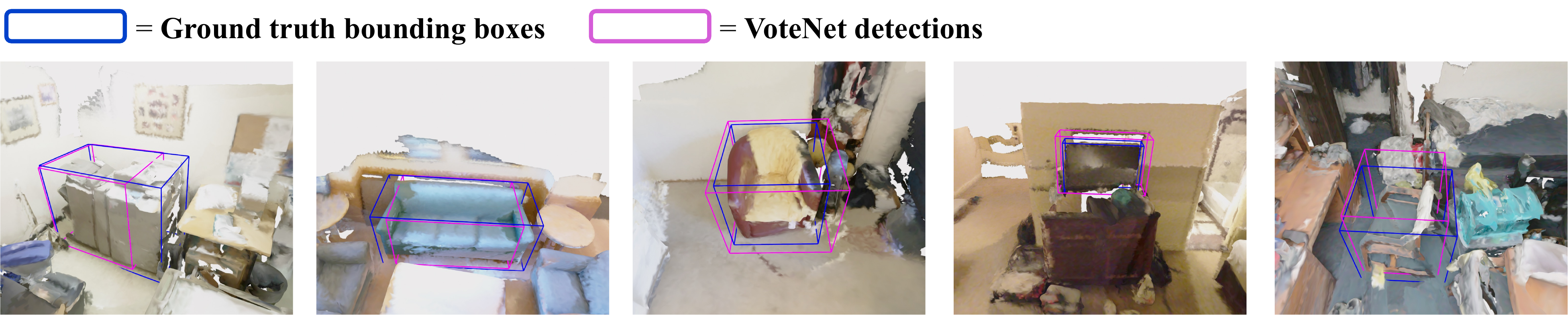}
  \caption{Comparison of ground truth bounding boxes (in blue) and VoteNet detections (in purple) used in the low guidance 3D visual grounding setting.}
  \vspace{128in}
\label{fig:app_votenet}
\end{figure*}

\paragraph{VoteNet detections}
In Figure~\ref{fig:app_votenet}, we show examples of VoteNet~\cite{qi2019deep} object detections, used in our low guidance setting, in comparison to ground truth bounding boxes. We see that VoteNet detections often result in incomplete point clouds, due to size corruption or center shift. This noise leads to additional challenges in 3D visual grounding; however, VoteNet detections significantly reduce the amount of labelled 3D data required during inference.

\section{Experiments}
\label{app:noise}

\paragraph{Noisy detection experiments.}
We report results of NS3D and \model over $6$ different levels of box prediction noise in Table~\ref{table:noise}, with each column representing ratio of perturbation on the original box. \model consistently improves NS3D under all settings.

\begin{table}[h]
\small
  \centering
  \begin{tabular}{lllllll}
    \toprule
        Noise level & $0.0$ & $0.1$ & $0.2$ & $0.3$ & $0.4$. & $0.5$ \\
    \midrule
    NS3D & $27.6$ & $22.9$ & $20.7$ & $19.6$ & $13.8$ & $10.7$ \\
    \model (Ours) & $\mathbf{36.6}$ & $\mathbf{35.6}$ & $\mathbf{33.5}$ & $\mathbf{30.2}$ & $\mathbf{24.7}$ & $\mathbf{20.1}$ \\
    \bottomrule
  \end{tabular}
  \caption{Comparisons under different levels of box prediction noise.}
  \label{table:noise}
  \vspace{-0.2cm}
\end{table}  
\section{ScanRefer}
\label{app:scanrefer}
Here, we describe how \model uses the ScanRefer~\cite{chen2020scanrefer} data for zero-shot transfer from ReferIt3D~\cite{achlioptas2020referit3d}.

\paragraph{Data construction}
We first create a subset of ScanRefer with queries that contain the same objects and relations as in ReferIt3D, such that we can run all method inference-only. This ScanRefer subset consists of $384$ unseen utterances, on the same ScanNet~\cite{dai2017scannet} scenes.

\paragraph{Implementation}
To transfer learned concepts to ScanRefer, we use GPT as \model's semantic parser to generate programs from input language. The programs are executed as described in the main paper. \model relies on the generalization abilities of LLMs to zero-shot transfer to ScanRefer, by decomposing new language into learned programs, without requiring any additional training or finetuning of neural networks. In comparison, end-to-end methods significantly underperform when faced with unseen input language.

\end{document}